# Facial age estimation using BSIF and LBP

Salah Eddine Bekhouche*, Abdelkrim Ouafi*, Abdelmalik Taleb-Ahmed** and Abdenour Hadid***, Azeddine Benlamoudi*

*Laboratory of LESIA, University of Biskra, Algeria, Email: salah@bekhouche.com, ou_karim@yahoo.fr, be.azzeddine@gmail.com
**LAMIH, UMR CNRS 8201 UVHC, University of Valenciennes, France, Email: taleb@univ-valenciennes.fr
*** Center for Machine Vision Research, PO Box 4500, FI-90014 University of Oulu, Finland, Email: hadid@ee.oulu.fi

*Abstract* – **Human face aging is irreversible process causing changes in human face characteristics such us hair whitening, muscles drop and wrinkles. Due to the importance of human face aging in biometrics systems, age estimation became an attractive area for researchers. This paper presents a novel method to estimate the age from face images, using binarized statistical image features (BSIF) and local binary patterns (LBP) histograms as features performed by support vector regression (SVR) and kernel ridge regression (KRR). We applied our method on FG-NET and PAL datasets. Our proposed method has shown superiority to that of the state-of-the-art methods when using the whole PAL database.**

## I. INTRODUCTION

Age estimation is a type of soft biometrics that provides ancillary information of an individual's identity information. It is defined as the age of a person based on their biometric features. It can be considered as either a multiclass classification problem or a regression problem [1], [2].

There are many popular real-world applications related to age estimation. For instance, internet access control, underage cigarette-vending machine use, security control and surveillance monitoring, biometrics ... etc.

A person's age can be determined in many ways but in our paper we focus on age estimation based on 2D face images of human subjects. Two main steps are used in age estimation systems, they are feature extraction and feature classification/regression. Feature extraction is essential since they affect the performance of the system. Feature classification/regression can be divided into three approaches: age group classification, single-level age estimation and hierarchical age estimation. Age group classification is an approach that predicts an age group. The single-level predicts an age label (value). The hierarchical age estimation is a hybrid approach i.e it is a combination of classification and regression methods [2].

## II. RELATED WORK

There are several studies in automatic facial age estimation. Those studies can be classified according to image representation and age estimation algorithms. Some studies are listed below.

Kwon and Lobo [3] presented a theory and practical computations to classify facial images into three groups: babies, young adults, and senior adults. The computations are based on cranio-facial development theory and skin wrinkle analysis.

Geng et al. [4], [5] proposed the aging pattern subspace. This theory is based on using a sequence of an individual's aging face images all together to model the aging process instead of dealing with each aging face image separately.

El Dib and El-Saban [6] presented a method of age estimation based on extended biologically inspired features (EBIF). Moreover, they combined regression-based and classification-based models.

Ricanek et al. [7] focuses on the development of a generalized multi-ethnic age-estimation technique. They used the active appearance model (AAM) due to its ability to capture relevant aging features.

Choi et al. [8] work is based on gaussian high pass filter (GHPF) for extracting local features and detailed age estimation which is performed by Support Vector Regression (SVR).

Nguyen et al. [9] re-defined the detected face region based on the distance between two eyes. They also applied the Multi-level local binary pattern (MLBP) method to extract the features for age estimation. In [10], they investigated the effects of gender and facial expression on age estimation using support vector regression (SVR) method.

## III. FACE AGING DATABASES

### A. FG-NET

The FG-NET[1] dataset contains 1002 images colored or gray with variation in resolution, quality, illumination, viewpoint and expression. There are 82 different subjects with ages ranging between newborns to 69 years old subjects. The average number of images per subject is 12 [11].

---
[1] http://www-prima.inrialpes.fr/FGnet/html/benchmarks.html

## B. PAL

The PAL[2] dataset contains 1046 face color images with variation in gender, ethnic and expression in the age range from 18 to 93 years old [12].

## IV. PROPOSED METHOD

### A. Overview

Our age estimation framework consists of three modules: the input image which assumes the existence of the face; the face pre-processing where the face is detected, normalized and cropped; the feature extraction where LBP and BSIF features of the face ROI are extracted; the age estimation which is a single-level age estimation using SVR and KRR and finally, the estimated age that is the output.

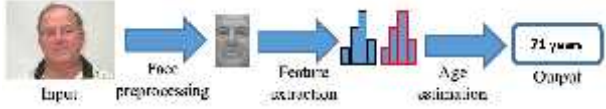

Fig. 1   Overview of the system

### B. Face pre-processing

We first convert all color face images into grayscale. Then we annotate each face using an active shape model with Stasm [13]. In the case of FG-NET we don't need to annotate the face because we have 68 landmarks defined by the popular FG-NET group.

In order to normalize the face, we use the center points of the eyes and rotate clockwise the face by an angle in (1), where the image center point $(C_x, C_y)$ is the center of rotation (Fig. 2).

$$\alpha = \tan^{-1}\left(\frac{R_y - L_y}{R_x - L_x}\right), \quad (1)$$

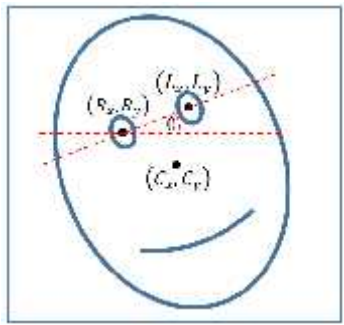

Fig. 2   Rotated face

After rotation, the new coordinates of the center points of the eyes are given by (2) and (3) (see Fig. 3).

$$R'_x = C_x + (R_x - C_x)\cos(\alpha) - (R_y - C_y)\sin(\alpha)$$
$$R'_y = C_y + (R_x - C_x)\sin(\alpha) + (R_y - C_y)\cos(\alpha), \quad (2)$$

$$L'_x = C_x + (L_x - C_x)\cos(\alpha) - (L_y - C_y)\sin(\alpha)$$
$$L'_y = C_y + (L_x - C_x)\sin(\alpha) + (L_y - C_y)\cos(\alpha), \quad (3)$$

[2]http://agingmind.utdallas.edu/facedb

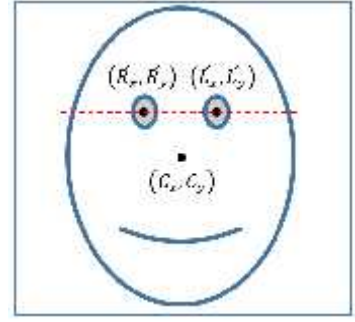

Fig. 3   Compensated face

The equations above are also applied on all landmarks: 68 landmarks in FG-NET and 78 in PAL databases. Finally, the face ROI is defined using the distance $l$ between the center points of the eyes (Fig. 4).

$$l = \sqrt{(R'_x - L'_x)^2 + (R'_y - L'_y)^2} = |R'_x - L'_x|, \quad (4)$$

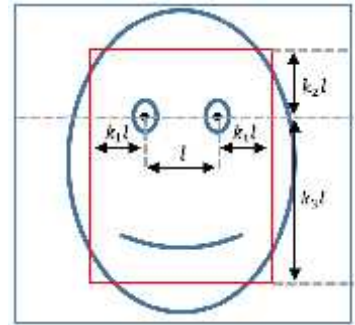

Fig. 4   Face ROI

Like in [9] the face ROI redefined using the $l$ value. The ratio values $k1$, $k2$ and $k3$ ($k1$=0.35, $k2$=1 and $k3$=1.75) was determined experimentally (Fig. 4). In the case of FG-NET database images, we define the face ROI used the min and the max landmarks (Fig. 5).

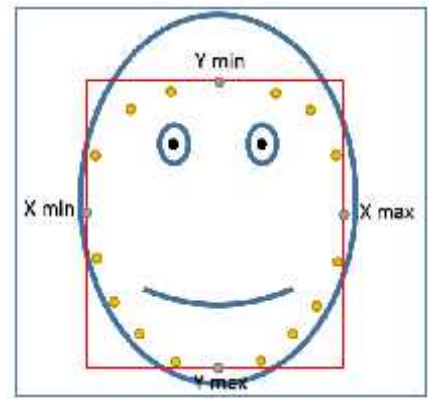

Fig. 5   Face ROI, FG-NET case

### C. Feature Extraction

In this paper, the face ROI features are extracted by LBP and BSIF methods. The LBP operator detects microstructures such as edges, spots and flat areas, it is one of the best performing texture descriptors and it also used in texture classification, segmentation, face detection, face recognition, gender classification, and age estimation applications [14], [15].

The original LBP operator works in a *3×3* neighborhood, each pixel can be labelled by using the center value as a threshold and considering the result as a binary number.

$LBP_{P,R}$ is almost used for pixel neighborhoods and it is refers to *P* sampling points on a circle of radius *R*.

The value of the LBP code of a pixel *(x$_c$, y$_c$)* is given by:

$$LBP_{P,R} = \sum_{p=0}^{P} s(g_p - g_c)2^p,  \quad (5)$$

where $g_c$ corresponds to the gray value of the center pixel *(x$_c$, y$_c$)*, $g_p$ refers to gray values of *P* equally spaced pixels on a circle of radius *R*, and *s* defines a thresholding function as follows:

$$s(x) = \begin{cases} 1 & if \quad x \geq 0 \\ 0 & otherwise. \end{cases} \quad (6)$$

Inspired by LBP and Local Phase Quantization (LPQ), Kannla and Rahtu [16] proposed a new local descriptor called BSIF (binarized Statisitcal Image features).

The basis vectors of a subspace into which local image patches are linearly projected are obtained from images by using Independent Component Analysis (ICA). The coordinates of each pixel are thresholded and thus a binary code is computed. The local descriptor of the image intensity patterns is represented by a value in the neighborhood of the considered pixel.

Given an image patch *X* of size *l×l* pixels and a linear filter $W_i$ of the same size, the filter response $s_i$ is obtained by:

$$s_i = \sum_{u,v} W_i(u,v) X(u,v) = w_i^T x, \quad (7)$$

where vectors *w* and *x* contain the pixels of $W_i$ and *X*.

The binarized feature $b_i$ is obtained by:

$$b_i = \begin{cases} 1 & if \quad s_i > 0 \\ 0 & otherwise. \end{cases} \quad (8)$$

The statistical independence of the filter responses is maximized to learn the set of filters from a training set of natural image patches via ICA.

In order to obtain the local and the global properties of image texture, we divided the face ROI into sub-blocks. Then we obtain the histogram features from each sub-block. Therefore, we get the whole image and 12 sub-block. We will have one vector consisting 13 histograms of BSIF features and the same for LBP. Finally we combine BSIF and LBP features to get 26 features (Fig. 6).

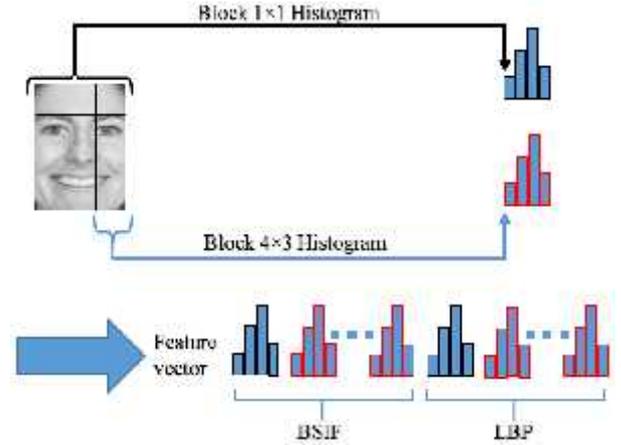
Fig. 6 Features extraction and combining

*D. Age Estimation*

Age estimation can be treated as a classification or regression problem. However, in our case, we treat it as regression problem so we used two regression algorithms with cross validations, which helps in selecting good parameters for the model. These algorithms are LIBSVM [17] for SVR and SimpleR [18] for KRR.

V. EXPERIMENTS

*A. Performance Evaluation*

To evaluate the performance of facial age estimation systems, there are two protocols **MAE** and **CS**.

In statistics, the **M**ean **A**bsolute **E**rror (MAE) is a quantity used to measure how close forecasts or predictions are to the eventual outcomes. The mean absolute error is given by (9) [19].

$$MAE = \frac{1}{n}\sum_{i=1}^{n}|f_i - y_i| = \frac{1}{n}\sum_{i=1}^{n}|e_i|, \quad (9)$$

- *n* : number of samples
- $f_i$ : the estimated value
- $y_i$ : the real value
- $e_i$ : error

The Cumulative Score (CS) can be viewed as an indicator of the accuracy of the age estimators. Since the acceptable error level is unlikely to be very high, the cumulative scores at lower error levels are more important. The cumulative score is given by (10) [5].

$$CS(l) = \frac{M_{e \leq l}}{M} 100\%, \quad (10)$$

- *l* : error level
- *M* : test images
- $M_{e\ l}$ : the number of test images on which the age estimation makes an absolute error no higher than *l* (years).

*B. Methodology*

We used PAL and FG-NET databases to evaluate the performance of different features extractor and regression algorithms. In PAL database the whole 1046 images was used, it divided into training and testing sets (523 images for each), CS curves are similarly shown in Fig. 7.

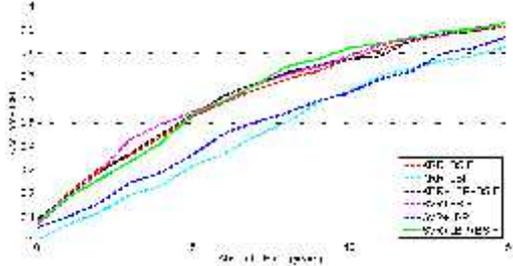

Fig. 7 Age estimation performance on the PAL

From the experiments, SVR+BSIF+LBP and KRR+BSIF+LBP give the smallest MAE, table I shows the comparison with the-state-of-the-art.

TABLE I
COMPARISONS IN PAL DATABASE

| METHODS OF AGE ESTIMATION | MAE | CS |
|---|---|---|
| SVR/SVM+LBP [8] | 8.44 | N/A |
| SVR+ MLBP [9] | 6.58 | N/A |
| SVR+ MLBP [10] | 6.52 | N/A |
| KRR+BSIF | 6.58 | 77% |
| KRR+LBP | 9.61 | 64% |
| KRR+BSIF+LBP | 6.38 | 77% |
| SVR+BSIF | 6.30 | 79% |
| SVR+LBP | 8.87 | 63% |
| SVR+BSIF+LBP | **6.25** | **82%** |

Due to the individual's age variation in FG-NET database, we followed the leave one person out (LOPO) strategy, each time a person's images are put into a test set whereas the other persons' images are put in a train set. CS curves are shown in Fig. 8.

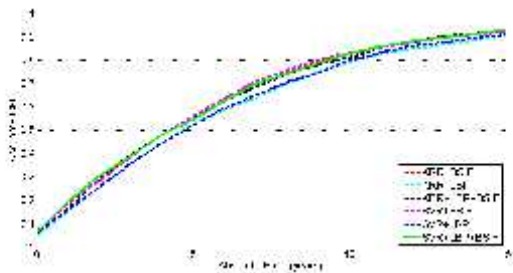

Fig. 8 Age estimation performance on the FG-NET.

From the experiments KRR+BSIF and SVR+BSIF+LBP give the smallest MAE, table II shows the comparison with the-state-of-the-art.

TABLE II
COMPARISONS IN FG-NET DATABASE

| METHODS OF AGE ESTIMATION | MAE | CS |
|---|---|---|
| WAS [20] | 8.06 | N/A |
| AGES [5] | 6.77 | 70% |
| QM [21] | 6.55 | N/A |
| $AGES_{lda}$ [5] | 6.22 | 81% |
| Bio-Inspired AAM [22] | 4.18 | 91% |
| Enhanced Bio-Inspired features [6] | 3.17 | 91% |
| KRR+BSIF | 6.29 | 83% |
| KRR+LBP | 7.05 | 83% |
| KRR+BSIF+LBP | 6.38 | 78% |
| SVR+BSIF | 6.32 | 83% |
| SVR+LBP | 6.95 | 80% |
| SVR+BSIF+LBP | 6.34 | 83% |

## VI. CONCLUSION

In this paper, we described a novel method for age estimation based on BSIF+LBP features and SVR regressor. The experimental results showed that the combination of BSIF and LBP features provide a better performance than the previous methods in the case of whole PAL database. In the future, we will try to redefine the face ROI taking into considering gender, ethnic and pose variation. We will also combine classification and regression to get a hybrid system.